\documentclass[10pt,twocolumn,letterpaper]{article}

\usepackage{cvpr}                 

\usepackage[table]{xcolor}
\usepackage[utf8]{inputenc}
\usepackage[T1]{fontenc}    
\usepackage{url}           
\usepackage{booktabs}     
\usepackage{amsfonts}     
\usepackage{nicefrac}      
\usepackage{microtype}    
\usepackage{xcolor}        
\usepackage{caption}     
\usepackage{lipsum}     
\usepackage{booktabs}
\usepackage[table]{xcolor} 
\usepackage{array}         
\usepackage{tabularx}
\newcolumntype{Y}{>{\raggedright\arraybackslash}X} 

\usepackage{graphicx}
\usepackage{float}
\usepackage{xspace}

\usepackage{amsmath} 
\usepackage{natbib}
\setcitestyle{numbers,square}
\usepackage{wrapfig}

\usepackage{multirow}
\definecolor{myblue}{RGB}{210, 225, 255}
\definecolor{logo}{HTML}{00a6fb}
\usepackage{dblfloatfix} 

\definecolor{cvprblue}{rgb}{0.21,0.49,0.74}
\usepackage[pagebackref,breaklinks,colorlinks,allcolors=cvprblue]{hyperref}

\title{RealGen: Photorealistic Text-to-Image Generation via Detector-Guided Rewards}

\author{
    Junyan Ye\textsuperscript{\rm 1,2}\textsuperscript{*}, 
    Leiqi Zhu\textsuperscript{\rm 1,3}\textsuperscript{*},
    Yuncheng Guo\textsuperscript{\rm 1}, 
    Dongzhi Jiang\textsuperscript{\rm 4},\\
    Zilong Huang\textsuperscript{\rm 2},
    Yifan Zhang\textsuperscript{\rm 5},
    Zhiyuan Yan\textsuperscript{\rm 6},
    Haohuan Fu\textsuperscript{\rm 5},
    Conghui He\textsuperscript{\rm 1},
    Weijia Li\textsuperscript{\rm 2}\textsuperscript{\dag} 
    \\
    \textsuperscript{\rm 1}Shanghai AI Lab,
    \textsuperscript{\rm 2}Sun Yat-Sen University, 
    \textsuperscript{\rm 3}Nanjing University,\\
    \textsuperscript{\rm 4}CUHK MMLab, 
    \textsuperscript{\rm 5}Tsinghua University,
    \textsuperscript{\rm 6}Peking University
}

\begin{document}

\twocolumn
[{
\renewcommand\twocolumn[1][]{#1}
\maketitle
\begin{center}
\vspace{-0.5cm}
\captionsetup{type=figure}
\includegraphics[width=0.95\linewidth]{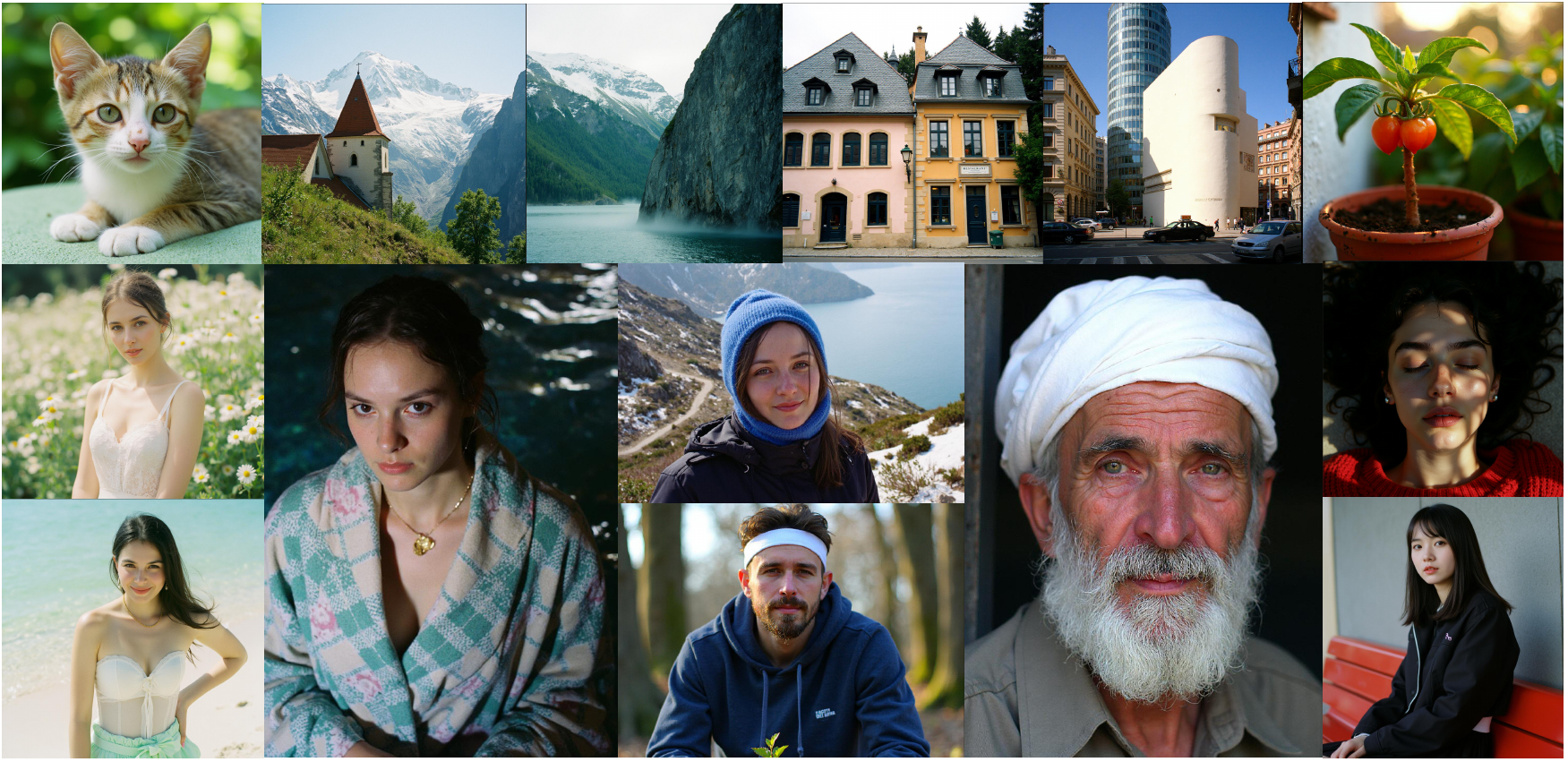}
\captionof{figure}{\textbf{Images generated by our proposed RealGen.} RealGen achieves superior photorealism and enhanced details, outperforming both powerful T2I models, such as Qwen-Image, and specialized photorealistic models, like FLUX-Krea.}
\label{fig:teaser}
\end{center}
}]

{
  \renewcommand{\thefootnote}{\fnsymbol{footnote}} 
  \footnotetext[1]{Equal contribution. \quad \textsuperscript{\dag}Corresponding author.} 
}

\begin{abstract} 

With the continuous advancement of image generation technology, advanced models such as GPT-Image-1 and Qwen-Image have achieved remarkable text-to-image consistency and world knowledge However, these models still fall short in photorealistic image generation. Even on simple T2I tasks, they tend to produce " fake" images with distinct AI artifacts, often characterized by "overly smooth skin" and "oily facial sheens". To recapture the original goal of "indistinguishable-from-reality" generation, we propose RealGen, a photorealistic text-to-image framework. RealGen integrates an LLM component for prompt optimization and a diffusion model for realistic image generation. Inspired by adversarial generation, RealGen introduces a "Detector Reward" mechanism, which quantifies artifacts and assesses realism using both semantic-level and feature-level synthetic image detectors. We leverage this reward signal with the GRPO algorithm to optimize the entire generation pipeline, significantly enhancing image realism and detail. Furthermore, we propose RealBench, an automated evaluation benchmark employing Detector-Scoring and Arena-Scoring. It enables human-free photorealism assessment, yielding results that are more accurate and aligned with real user experience. Experiments demonstrate that RealGen significantly outperforms general models like GPT-Image-1 and Qwen-Image, as well as specialized photorealistic models like FLUX-Krea, in terms of realism, detail, and aesthetics. The code is available at \url{https://github.com/yejy53/RealGen}.

\end{abstract}

\section{Introduction}

\begin{figure}[t]
    \centering
    \includegraphics[width=0.95\linewidth]{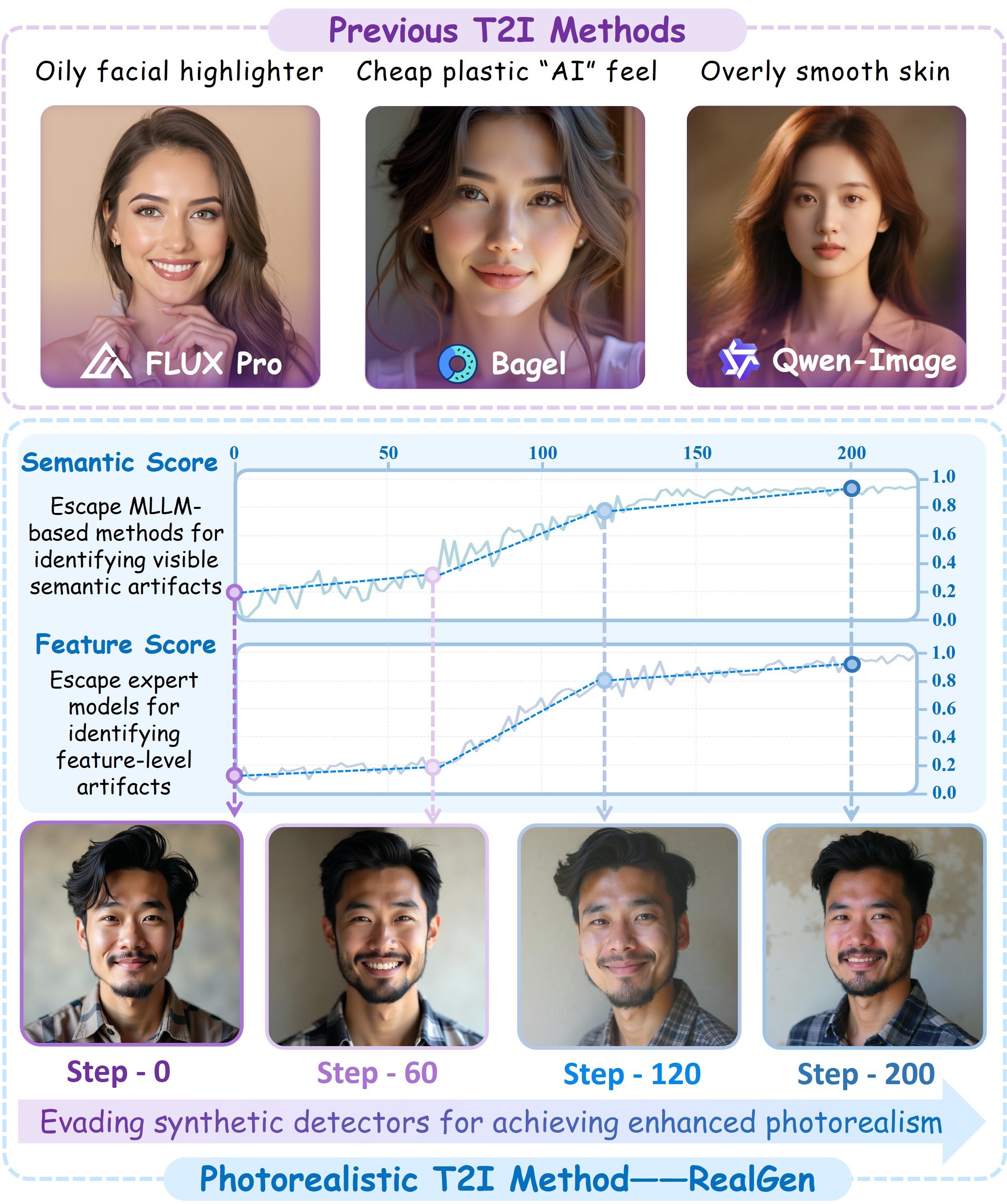}
    \vspace{-10pt}
    \caption{\textbf{From Synthetic Artifacts to Photorealism.} Contrasting the common "fake-feel" AI artifacts in previous T2I methods , our proposed RealGen achieves enhanced photorealism by progressively evading semantic and feature-level detectors.}
    \label{fig:2}
    \vspace{-15pt}
\end{figure}

Image generation has undergone a significant evolution from GANs~\cite{goodfellow2014generative} to Diffusion models~\cite{rombach2022high,sd3}, leading to a generation of powerful models such as GPT-Image-1~\cite{gpt4o}, Nano-Banana~\cite{google2025gemini}, and Qwen-Image~\cite{wu2025qwen}. These models demonstrate exceptional capabilities in areas like precise attribute control and complex text rendering. However, recent research has focused on enhancing prompt fidelity and leveraging world knowledge for complex content generation. Despite these advancements, even existing powerful generative models like FLUX.1 Pro~\cite{flux2024} and Qwen-Image~\cite{wu2025qwen} still tend to produce images that lack photorealism, such as "overly smooth skin" and "oily facial sheens", as shown in Fig. \ref{fig:2}. This arguably deviates from the original goal of image generation: \textit{\textbf{"to produce visuals that are indistinguishable from reality,"}} which is precisely the focus of our work.

To pursue higher photorealism, existing research has attempted to combine generative models with reinforcement learning (RL) and human preference scores~\cite{wallace2024diffusion,shen2025directly}. For instance, methods like DanceGRPO~\cite{xue2025dancegrpo} and FlowGRPO~\cite{liu2025flow} utilize reward models such as PickScore~\cite{kirstain2023pick} and HPSv2~\cite{wu2023human} to align generation with human preferences. However, trained on human preference data may introduce prior biases, such as an excessive preference for colors like red or purple. More critically, human preference scores do not directly equate to photorealism. A candid snapshot might be highly realistic but receive a low score due to a lack of aesthetic appeal, inadvertently encouraging the model to favor artistic or anime styles. Meanwhile, approaches like FLUX-Krea~\cite{flux1kreadev2025} employ large-scale, human-curated, high-quality image samples combined with TPO to enhance realism. However, this method is not only cost-prohibitive but also highly dependent on the subjective preferences of annotators. Therefore, the critical bottleneck in leveraging RL for photorealism lies in \textit{how to establish an objective, scalable, and human-free metric for image realism.}

Inspired by adversarial generation, a promising approach is to quantify photorealism using synthetic image detectors. The core rationale is that\textbf{\textit{ a more realistic image should possess fewer AI artifacts and thus be more difficult for a detector to classify as "Fake."}} In recent years, as generative models have rapidly advanced, corresponding image detectors have evolved in parallel, achieving robust detection performance. Furthermore, some MLLM-based methods have extended this capability to semantic-level, interpretable detection, such as analyzing "AI-feel" skin textures. Therefore, we posit that two distinct classes of detectors can be employed to measure image realism: one for analyzing visible, semantic-level artifacts, and another for deep feature-level artifacts. As shown in Figure \ref{fig:2}, during the RL process, we define these detectors as the reward model to guide the generator to "escape" detection, which effectively reduces artifacts and enhances image realism.

In addition, the quality of image synthesis is highly correlated with the complexity of the text prompt~\cite{cao2023beautifulprompt}. Expert users, through sophisticated "prompt engineering," can produce photorealistic, cinematic-quality images. Conversely, simple prompts from casual users often yield low-quality results that lack realism. This discrepancy arises because T2I models often rely on complex prompt structures; simple prompts provide low information entropy, causing the model to default to its high-frequency priors, which results in images lacking detail and specificity~\cite{zhang2025intricate}. Therefore, automatically rewriting or enriching the user's input directive presents another critical pathway to enhancing the quality and realism of the subsequently generated images~\cite{wang2025promptenhancer}.

In summary, we propose \textbf{\textit{RealGen}}, a framework for photorealistic text-to-image generation. The framework includes a Large Language Model (LLM) component for optimizing user prompts and a diffusion model dedicated to generating realistic images. We utilize detector models as rewards, optimizing both the LLM and the generative model via Generalized Reinforcement Policy Optimization (GRPO) algorithm~\cite{liu2025flow,xue2025dancegrpo} to effectively reduce artifacts in the generated images. Specifically: (1) With the generative model held fixed, we train the LLM to optimize the input prompts, using the detector model to score the final synthesized image; this encourages the LLM to generate richer and more effective prompts. (2) While maintaining text input consistency, we optimize the diffusion model itself, enabling it to "escape" the detectors and generate more realistic and detailed images.

Furthermore, we also propose \textbf{\textit{RealBench}}, a new benchmark for evaluating the photorealism of generated images. RealBench contains a diverse set of text prompts, supplemented by high-quality, real-world image references. Addressing the lack of effective evaluation methods beyond manual human assessment, RealBench employs two automated evaluation protocols: Detector-Scoring and Arena-Scoring. First, Detector-Scoring utilizes multiple synthetic image detectors (including those held out from the reward) to score the images; images with fewer AI artifacts receive higher scores. Second, for Arena-Scoring, we adapt the LMArena methodology, using several different MLLMs to simulate user preferences in pairwise comparisons, selecting the result that appears more realistic. Outputs from each model are put into pairwise "battles" against other models or real-world images for at least 3000 random pairings to determine a final win rate. The inclusion of real-world data not only enhances scoring stability but also validates the effectiveness of this adversarial scoring approach. Our main contributions are as follows:

\begin{itemize}
  \item We propose \textbf{\textit{RealGen}}, a text-to-image generator capable of producing highly convincing photorealistic images.

It leverages a Detector Reward-guided GRPO post-training to escape detector identification, thereby reducing artifacts and enhancing image realism and detail.

  \item We introduce \textbf{\textit{RealBench}}, a new benchmark for evaluating photorealism that achieves human-free automated scoring through Detector-Scoring and Arena-Scoring.

  \item RealGen significantly outperforms both general image models (like GPT-Image-1, Qwen-Image) and specialized realistic models (like FLUX-Krea) in realism, details, and aesthetics on the T2I task.
\end{itemize}
\section{Related Work}

\subsection{Image Generation Models}

Image generation models have made significant progress in recent years~\cite{chen2023pixart,yan2025gpt,brooks2023instructpix2pix,deng2024nova}, demonstrating exceptional performance across diverse downstream tasks~\cite{li2024crossviewdiff,ye2025satellite,ye2024skydiffusion}.
Representative models such as such as Stable Diffusion ~\cite{LDM,podell2023sdxl,sd3}, FLUX~\cite{flux2024,fluxkontext}, Emu~\cite{sheynin2024emu,cui2025emu3,wang2024emu3} and DALL·E~\cite{dalle2} demonstrating powerful text-to-image generation capabilities. As research gradually shifts toward multimodal generative models, these models achieve understanding and generation through unified architectures~\cite{xie2024show,pan2025metaquery,yan2025can,chen2025blip3}. Although powerful models like GPT-Image-1~\cite{gpt4o}, Bagel~\cite{bagel} and OmniGen2~\cite{wu2025omnigen2} can generate precise objects and complex text, challenges remain in generating realistic images, especially with the strong oily appearance of human faces. RealGen focuses on enhancing realistic image generation, optimizing image realism and detail.

\subsection{Reinforcement Learning and Human Preferences in Image Generation}

Recently, with the development of reinforcement learning, more methods have begun to explore its application in the Diffusion image generation domain~\cite{xue2025dancegrpo,yan2025can}. For example, DiffusionDPO~\cite{wallace2024diffusion} and FlowGRPO~\cite{liu2025flow} use human preference data as a reward function to generate images that align with human aesthetics. However, reward models based on human preferences can introduce prior biases, such as HPSv2~\cite{wu2023human} favoring red-toned images and PickScore~\cite{kirstain2023pick} preferring purple ones. While human-captured images are generally realistic, they do not always score highly in terms of preference. SRPO~\cite{shen2025directly} has made some progress in enhancing image realism but still lacks in aesthetic quality. FLUX-Krea~\cite{flux1kreadev2025}, through large-scale manual collection of high-quality image samples and TPO reinforcement learning, optimizes the realism of generated images. However, this method is limited by the costs of data labeling and the personal preferences of annotators. In contrast, RealGen does not rely on human preference data. Instead, it uses advanced AIGC detection models to guide the generative model away from AI artifacts at both semantic and feature levels, further improving image realism.

\subsection{Synthesis detection for Image generation}

As the realism of image generation continuously increases, corresponding detection techniques have rapidly evolved~\cite{ye2024loki,lin2025seeing}. From early detectors like CNNSpot~\cite{wang2020cnn} advanced methods leveraging powerful pre-trained vision models (e.g., OmniAID~\cite{guo2025omniaid}, Effort~\cite{yan2024effort}), these detectors have demonstrated success rates exceeding 80\% ~\cite{zhu2023genimage,ye2024loki}.
Recently, detector based on MLLMs have emerged, such as FakeVLM~\cite{wen2025spot} and LEGION~\cite{kang2025legion}, which not only achieve good detection performance but also enhance the explainability of artifact detection. Since synthetic image detection and generation are a "cat-and-mouse" game, some studies have explored using detector feedback to optimize generation quality. However, these methods are currently more limited to post-processing techniques, such as Inpainting, to optimize already generated images~\cite{zhang2022perceptual,zhang2023perceptual}. In contrast, our work combines reinforcement learning and synthetic detectors to directly optimize the generative model itself.

\section{RealGen}
\begin{figure*}[ht]
    \centering
    \includegraphics[width=1\linewidth]{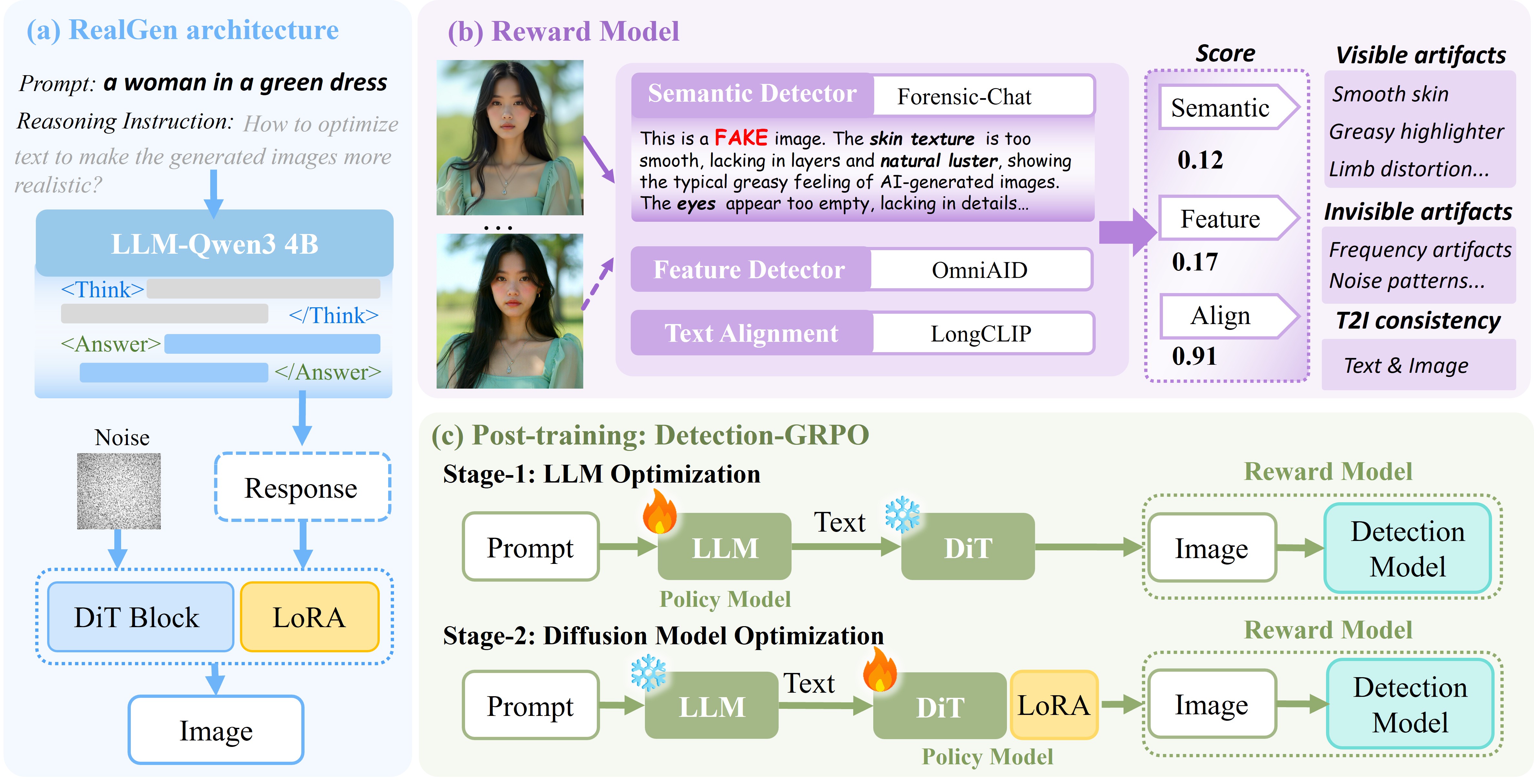}
    \caption{\textbf{Overview of the RealGen Method.} (a) The architecture of RealGen, consisting of an LLM component and a Diffusion component. (b) Our detector-based reward model, which evaluates images based on visible artifacts, feature-level artifacts, and text-image alignment. (c) The two-stage post-training process guided by this reward model, which respectively optimizes the LLM and Diffusion components.}
    \label{fig:3}
\end{figure*}

\subsection{Model architecture}

As illustrated in Figure \ref{fig:3}, the architecture of RealGen comprises two core components: a LLM for understanding and refining user intent, and a diffusion model for realistic image synthesis. First, the LLM receives the initial user instruction and performs "thought and planning": it expands the short prompt into a longer, more diverse text description by adding rich details. Subsequently, the diffusion model utilizes this refined text as its input condition, executing the denoising and decoding process to generate the final image.

For implementation, we employ Qwen-3 4B~\cite{yang2025qwen3} as the base LLM. For image generation, we utilize the advanced pre-trained diffusion model, FLUX.1-dev dev~\cite{flux2024}, integrated with fine-tuned LoRA layers. Both the LLM and the diffusion model first undergo specialized Supervised Fine-Tuning (SFT) as a cold-start phase, followed by Reinforcement Learning optimization via GRPO. The subsequent sections will detail our designed detection reward function and the RL training process.

\subsection{Detection Reward}

To steer the model optimization towards high-fidelity realism, the design of the reward function is critical, as it must accurately quantify authenticity. We adopt a "detection-as-reward" paradigm inspired by adversarial generation, designing a multi-objective reward function. As shown in Fig. \ref{fig:3}(b), this function combines detectors at two distinct levels—semantic and feature—to penalize both perceptible artifacts and imperceptible synthesis traces, respectively.

\textbf{Semantic Detector:} We employ Forensic-Chat~\cite{lin2025seeing}, a generalizable and interpretable detector optimized from Qwen2.5-VL-7B~\cite{wang2024qwen2}. It assesses authenticity by analyzing image content (e.g., smooth greasy skin, artifacts in faces/hands, unnatural background blur). We define the semantic reward $R_{\text{semantic}}$ as the normalized probability of its output "real" token probability:
\begin{equation}
R_{\text{semantic}} = \text{softmax}\left( \left[ L_{\text{("fake","Fake")}}, L_{\text{("real","Real")}} \right] \right)_1
\end{equation}

\textbf{Feature Detector:} We utilize the advanced expert detector OmniAID~\cite{guo2025omniaid}, which achieves stable and accurate detection by being pre-trained on large-scale real and synthetic datasets. Feature-level artifacts are primarily associated with frequency artifacts and abnormal noise patterns. We define the feature-level reward ($R_{\text{feature}}$) as one minus its output "fake" probability, i.e., $1 - P_{\text{OmniAID}}(\text{fake})$.

\textbf{Text Alignment:} Finally, we include the Long-CLIP~\cite{zhang2024long} score as an auxiliary reward ($R_{\text{align}}$) to maintain text-image alignment. This prevents the model from sacrificing fidelity to the input prompt in its pursuit of realism.

For the GRPO process, we combine the three reward functions defined above: $\{R_{\text{semantic}}, R_{\text{feature}}, R_{\text{align}}\}$. For a batch of $N$ samples $\{ I_i \}_{i=1}^N$, we first evaluate each sample $I_i$ to obtain its raw scores $\{r_i^{\text{sem}}, r_i^{\text{feat}}, r_i^{\text{align}}\}$. Since each reward function operates on a different scale and distribution, we first normalize the scores for each reward dimension within the batch. Then, we sum these normalized scores to fuse them into a single advantage function $A(I_i)$, as defined in Equation (\ref{eq:advantage}):

\begin{equation}
A(I_i) = \sum_{k \in \{\text{sem, feat, align}\}} \frac{r_i^k - \text{Mean}(\{r_j^k\}_{j=1}^N)}{\text{Std}(\{r_j^k\}_{j=1}^N)}
\label{eq:advantage}
\end{equation}

\subsection{Post-training: Detection-GRPO} 

\textbf{LLM Optimization.}
In the first stage, we optimize the LLM for user intent refinement, while the parameters of the image generation model remain frozen. In this setup, the LLM $\pi_\theta$ acts as the policy network. We first conduct SFT as a cold-start to teach it the "think-plan-then-generate" pattern required by the system prompt. During the RL phase, for a given text input $x$, the policy LLM samples and rewrites it, generating $N$ optimized prompts $\{ y_i \}_{i=1}^N$. Subsequently, the frozen diffusion model generates an image $I$ conditioned on $y$. This image $I$ is then evaluated by our multi-objective reward function to obtain the reward $R(I, y)$. In this context, each trajectory $\{ o_i \}_{i=1}^N$ corresponds to the text sequence $y_i = (y_{1,i}, \dots, y_{T,i})$ sampled by the LLM. We update the LLM's parameters by maximizing the GRPO objective function defined in Equation (\ref{eq:GRPO}).
\begin{multline}
\mathcal{J}(\theta) = \mathbf{E}_{c \sim \mathcal{C}, \{o_i\}_{i=1}^N \sim \pi_{\theta_{\text{old}}}(\cdot|c)} \\ 
\Biggl[ \frac{1}{N} \sum_{i=1}^N \frac{1}{T} \sum_{t=1}^{T} \biggl( \mathcal{J}_{GRPO} - \beta \mathbf{D}_{\text{KL}}(\pi_\theta \| \pi_{\theta_{\text{ref}}}) \biggr) \Biggr], 
\label{eq:GRPO}
\end{multline}
$$
\mathcal{J}_{GRPO} = \min \left( \rho_{t,i} A_i, \text{clip}(\rho_{t,i}, 1 - \varepsilon, 1 + \varepsilon) A_i \right)
$$

where $A_i$ is the fused advantage function obtained from Equation (\ref{eq:advantage}) and $\rho_{t,i}$ denotes the importance sampling ratio between policies, which is defined in Equation (\ref{eq:policy1}). 
\begin{equation}
\rho_{t,i} = \frac{\pi_\theta(y_{t,i} | x, y_{<t,i})}{\pi_{\theta_{\text{old}}}(y_{t,i} | x, y_{<t,i})}
\label{eq:policy1}
\end{equation}
The optimized LLM learns to explore and generate superior prompts, for instance: (1) adding rich scenic details, (2) naturally incorporating "flaws and imperfections," and (3) appending specific auxiliary words (e.g., "shot on iPhone") that aid the diffusion model in producing realistic images.

\textbf{Diffusion Model Optimization.} In the second stage, we optimize the diffusion model for photorealistic image generation, while keeping the parameters of the LLM component frozen. We employed a diffusion model RL framework Flow-GRPO~\cite{liu2025flow}, however, due to the stochastic nature of RL, images generated through limited denoising steps or full-trajectory exploration tend to be noisy and blurry~\cite{shen2025directly}. This unstable process subsequently misleads the judgment of detection models, which are typically trained on datasets lacking exposure to such noisy or ambiguous artifacts. To address this training-inference inconsistency, for a given input $x$, the policy model $p_\theta$ executes a complete denoising inference $(x_T, x_{T-1}, \dots, x_0)$ for evaluation. Simultaneously, several consecutive steps $\Delta t$ from this process is designated for random exploration and obtain $N$ trajectories $(x_{T^\prime,i}, x_{T^\prime-1,i}, \dots, x_{T^\prime-\Delta t,i})$, which effectively balance RL explorativeness with training efficiency. We optimize the policy model through GRPO process the same as stage-1, as defined in Equation (\ref{eq:policy2}), and the probability ratio $\rho_{t,i}$ is given by:
\begin{equation}
\rho_{t,i} = \frac{p_\theta(x_{t-1,i} | x_{t,i}, c)}{p_{\theta_{\text{old}}}(x_{t-1,i} | x_{t,i}, c)}
\label{eq:policy2}
\end{equation}
SDE sampling provides stochasticity to the reverse process. During GRPO training, a portion of short texts are processed by the text-rewrite LLM, while the rest bypass this component. This strategy enhances the model's generalization robustness across prompts of varying lengths. Finally, detector reward optimization enables diffusion models to significantly reduce semantic and feature-level artifacts, consequently generating more realistic images.
\section{RealBench}
\subsection{Dataset overview}

\begin{figure}[t]
    \centering
    \includegraphics[width=\linewidth]{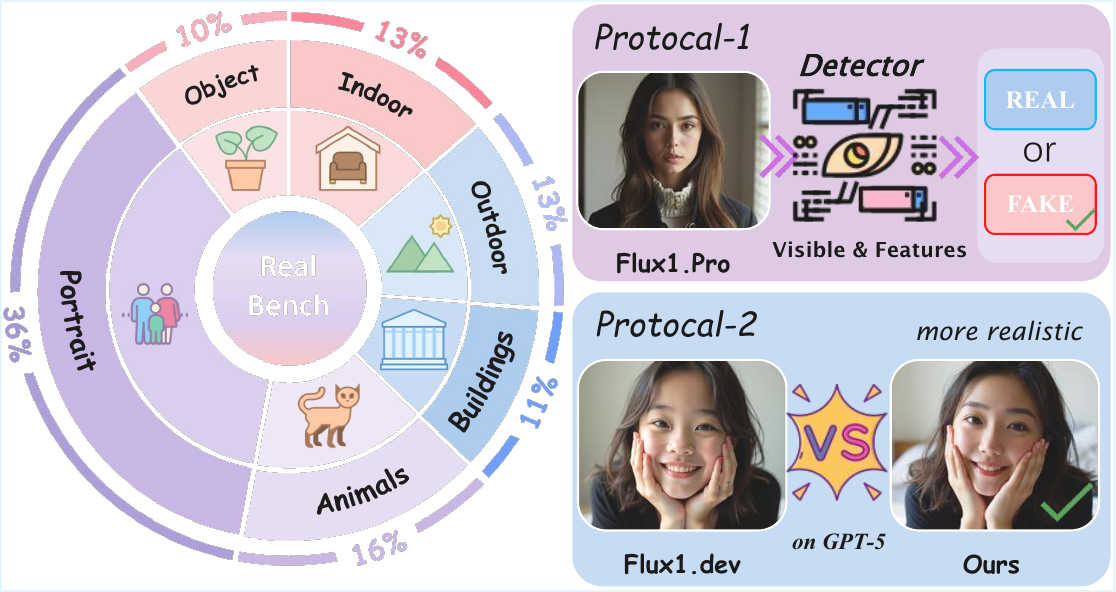}
    \vspace{-5pt}
    \caption{\textbf{Overview of the RealBench.} The left shows the categorical data composition. The right details its evaluation protocol.}
    \label{fig:Benchmark}
    \vspace{-10pt}
\end{figure}

We introduce RealBench, a benchmark platform designed to comprehensively evaluate the photorealism of T2I synthesized images. As illustrated in Fig. \ref{fig:Benchmark}, RealBench features a meticulously curated dataset of 1000 high-quality, real-world images and their corresponding captions, sourced from the internet and free photography websites \footnote{https://pixabay.com/;    https://www.pexels.com/}. This dataset spans seven distinct categories. Recognizing that "Portrait" is one of the most common and challenging categories in user T2I prompts, we have significantly increased its proportion while ensuring diversity across other categories. Unlike existing benchmarks focusing on instruction following (e.g., GenEval~\cite{ghosh2023geneval}, DPG-Bench~\cite{hu2024ella}) or human preference (e.g., HPD V2~\cite{wu2023humanpreferencescorev2}), RealBench focuses exclusively on assessing the photorealism of generated results. RealBench comprises two key evaluation protocols: Protocol-1, detector-based realism quantification (Detector-Scoring); and Protocol-2, arena-style preference evaluation (Arena-Scoring).

\begin{table*}[t]
    \centering
    \renewcommand{\arraystretch}{1.2}
    \caption{Evaluation of image generation ability on RealBench. \textbf{Bold} indicates the best result, and \underline{underlined} denotes the second best. * indicates that we used an LLM component for prompt optimization.}
    \label{tab:1}
    \vspace{-5pt}
    \resizebox{\textwidth}{!}{
        \begin{tabular}{lcccccccccc} 
            \toprule
            \multirow{2}{*}{\textbf{Model}} & \multicolumn{4}{c}{\textbf{Detector-Scoring}} & \multicolumn{2}{c}{\textbf{Arena-Scoring}} & \multicolumn{4}{c}{\textbf{Other metrics}} \\ 
            
            \cmidrule(lr){2-5} \cmidrule(lr){6-7} \cmidrule(lr){8-11} 
            
            & Forensic-chat & OmniAID & Effort & GPT 5-Prompt & VS Real & VS Others & PickScore & CLIP & HPSv2.1 & HPSv3 \\ 
            
            \midrule
            
            \multicolumn{11}{l}{\textit{Closed-Sourced T2I Model}} \\ 
            \midrule
            \quad FLUX-Pro~\cite{flux2024} & 57.45 & 21.55 & \textbf{20.94} & 50.14 & 18.20 & - & 23.68 & 86.85 & 30.79 & \underline{12.78} \\
            \quad Nano-Banana~\cite{google2025gemini} & 46.43 & \underline{31.02} & \underline{11.74} & \underline{73.19} & \textbf{42.17} & - & 23.67 & 84.02& 30.90 & \textbf{12.95} \\ 
            \quad SeedDream 3.0~\cite{flux2024} & \underline{63.47} & 23.73 & 8.91 & \textbf{79.92} & \underline{36.40} & - & \textbf{23.92} & \textbf{88.61} & \textbf{31.48} & 12.02 \\  
            \quad GPT-Image-1~\cite{gpt4o} & \textbf{75.63} & \textbf{33.59} & 5.70 & 70.98 & 33.71 & - & \underline{23.89} & \underline{88.57} & \underline{31.25} & 12.48 \\ 
            
            \midrule
            
            \multicolumn{11}{l}{\textit{Open-Sourced T2I Model}} \\ 
            \midrule

            \quad SDXL~\cite{podell2023sdxl} & 43.37 & 24.44 & 8.44 & 23.82 & 9.22 & 35.20 & 23.02 & 84.65 & 28.44 & 9.87 \\ 
            \quad SD-3.5-Large~\cite{sd3} & 48.63 & 20.02 & 17.45 & 76.46 & 23.82 & 55.58 & 23.46 & \underline{88.23} & 30.68 & 12.06 \\ 
            \quad FLUX.1-dev~\cite{flux2024} & 40.91 & 21.32 & 14.85 & 43.03 & 12.61 & 43.60& \underline{23.59} & 86.33 & \underline{31.21} & \underline{13.58} \\ 
            \quad FLUX.1-Kontext~\cite{fluxkontext} & 37.20 & 20.68 & 10.68 & 10.76 & 3.65 & 20.40 & 22.80 & 84.20& 30.08 & 11.61 \\
            \quad Echo-4o~\cite{ye2025echo} & 38.86 & 16.93 & 15.71 & 17.86 & 6.34 & 34.35 & 23.53& \textbf{89.25}& 31.69 & 12.27 \\ 
            \quad Bagel~\cite{bagel} & 39.47 & 17.77 & 14.47 & 21.32 & 5.47 & 22.95 & 23.39 & 87.92 & 31.19 & 12.43 \\
            \quad Qwen-Image~\cite{wu2025qwen} & 57.47 & 36.82 & 17.10 & 65.03 & 18.25 & 47.35 & 21.97 & 82.83 & 25.50 & 8.15 \\ 
            
            \quad SRPO~\cite{shen2025directly} & 64.14 & \underline{40.09} & 24.73 & 81.29 & 40.70 & 64.30 & 23.55& 86.16 & 29.66 & 12.43 \\ 
            \quad FLUX-Krea~\cite{flux1kreadev2025} & 57.10 & 32.44 & 18.42 & 79.44 & 37.60 & 66.40 & \textbf{23.77} & 87.60 & 30.75 & 12.50 \\ 
            \rowcolor{myblue}
            \quad \textbf{Ours} & \underline{70.59} & 37.85 & \underline{31.71} & \underline{92.79} & \underline{43.41} & \underline{74.80} & 23.58 & 86.80 & \textbf{31.87} & \textbf{13.61} \\  
            \rowcolor{myblue}
            \quad \textbf{Ours*} & \textbf{80.84} & \textbf{47.20 }& \textbf{38.35} & \textbf{96.73} & \textbf{50.15} & \textbf{84.85} & 21.75 & 87.69 & 28.24 & 11.11 \\ 
            
            \bottomrule
        \end{tabular}
    }
    \vspace{-15pt}
\end{table*}

\subsection{Detector-Based Realism Quantification}

The core motivation for this protocol is that more photorealistic images should better evade advanced synthetic detectors, thus lowering their probability of being classified as "fake." Therefore, we utilize the probability of an image being deemed "real" as its photorealism score. In our assessment, we deploy detectors at both the visible semantic layer and the invisible feature layer. First, we include Forensic-Chat~\cite{lin2025seeing} and OmniAID~\cite{guo2025omniaid}, which are consistent with our reward function. At the semantic level, we also employ a leading closed-source MLLM (GPT-5~\cite{openai_gpt5_system_card_2025}) as a discriminator, guided by strict prompts (see supplementary material) to focus on common artifact regions. At the feature layer, we additionally use Effort~\cite{yan2024effort}, an unrelated expert synthetic detector known for its strong generalization in assessing image authenticity. The inclusion of these two reward-independent detectors helps ensure the comprehensiveness and robustness of the evaluation.

\subsection{Arena-Style Preference Evaluation}

We observe that it is inherently difficult for preference models, expert models, or even MLLMs to provide reliable, absolute quantitative scores for a single image. Therefore, inspired by LM-Arena~\footnote{https://lmarena.ai/leaderboard/text-to-image}, we adopt a pairwise comparison protocol, termed Arena-Scoring. In this protocol, we employ GPT-5 as the "judge model" to simulate user preferences. During evaluation, the judge model is presented with two semantically similar images corresponding to the identical text prompt(e.g., Model A's output vs. Model B's output, or Model A's output vs. a real image) and is compelled to make a forced-choice decision, selecting the one it perceives as more realistic. Finally, the images generated by each model undergo at least 3000 random pairwise "battles" against outputs from other models and real-world data. We then calculate the final win rate. The inclusion of real images in the comparison pool not only enhances scoring stability but also allows us to validate the plausibility and reliability of the MLLM judge's preferences.
\section{Experiments}
\label{sec:experiment}

\begin{figure*}[t]
    \centering
    \includegraphics[width=0.97\linewidth]{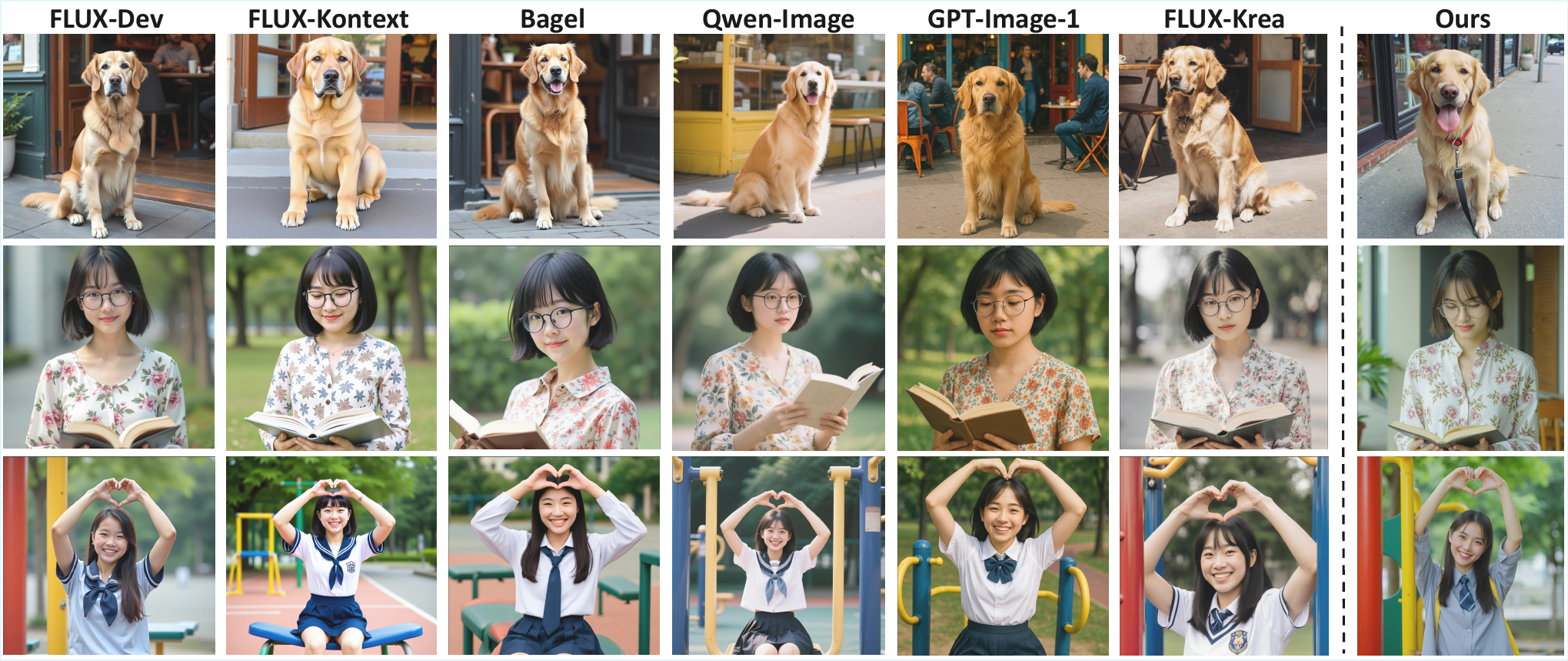}
    \vspace{-5pt}
    \caption{Qualitative comparison of different methods on RealBench.}
    \label{fig:5}
    \vspace{-10pt}
\end{figure*}

\subsection{Experimental Setup}
\label{subsec:t2i}

\textbf{Implement Details.} Our RealGen model employs Qwen3-4B-Instruct~\cite{yang2025qwen3} as the LLM component and FLUX.1.dev~\cite{flux2024} as the base image generator. The training data for SFT and RL is primarily sourced from the real image subset of HPD v3~\cite{ma2025hpsv3}. All experiments are conducted on 8 H200 GPUs. For the first stage of GRPO, we set the batch size to 32; for the second stage, the batch size is 12. The entire RL training process iterates for approximately 230 steps. To ensure a fair evaluation, we assessed model effectiveness on two datasets independent of the HPD v3 training data: our newly proposed RealBench and the "Photo" subset of HPD v2~\cite{wu2023humanpreferencescorev2}.

\noindent \textbf{Comparison Method.} We conduct extensive comparisons between RealGen and current T2I models. These baselines include general generative models, such as the closed-source Flux.1 Pro~\cite{flux2024}, GPT-Image-1~\cite{gpt4o}, and Nano Banana~\cite{google2025gemini}, as well as open-source models like Flux.1 dev~\cite{flux2024}, Bagel~\cite{bagel}, and Qwen-Image~\cite{wu2025qwen}. For a fair comparison, we also benchmark against methods specifically employing RL optimization for human preference or realism, such as Flux-Krea~\cite{flux1kreadev2025} and SRPO~\cite{shen2025directly}. To ensure a fair comparison, all baselines utilize their official default settings.

\noindent \textbf{Evaluation Protocol.} 
We employ a comprehensive evaluation protocol composed of three main categories of metrics: (1) Detector-Scoring, (2) Arena-Scoring, and (3) Other Metrics. The detailed configurations for the first two categories are described in Section 4 (RealBench). The third category, Other Metrics, includes preference alignment scores such as Pick-Score~\cite{kirstain2023pick}, HPSv2.1~\cite{wu2023humanpreferencescorev2}, and HPSv3~\cite{ma2025hpsv3}, as well as a LongCLIP~\cite{zhang2024long} image-text alignment metric. It is worth noting that HPSv3, unlike its predecessors, incorporates real-world images into its training data, allowing it to more comprehensively reflect human preferences for both photorealism and overall quality.

\subsection{Main Results}

\begin{figure}[t]
    \centering
    \includegraphics[width=0.95\linewidth]{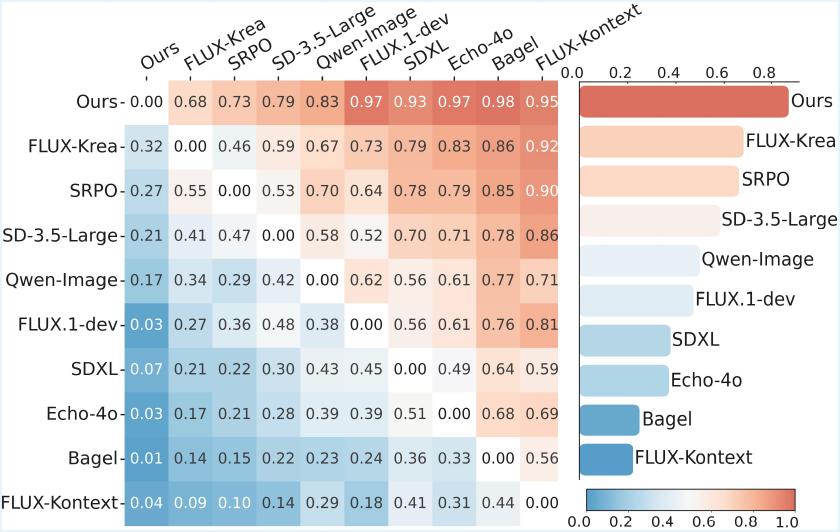}
    \vspace{-10pt}
    \caption{Pairwise Realism Comparison Matrix of Open-Sourced Text-to-Image Models.}
    \label{fig:6}
    \vspace{-5pt}
\end{figure}

Table~\ref{tab:1} presents the evaluation results of different methods on RealBench. Our proposed RealGen outperforms existing leading T2I models across multiple key photorealism metrics, regardless of whether the LLM prompt rewrite component is used. For instance, images generated by general-purpose open-source T2I models, such as FLUX.1-kontext and Bagel, are easily identified by synthetic detectors, exposing obvious AI artifacts. Methods specialized for photorealism, like Flux-Krea and SRPO, achieve sub-optimal performance, but still a significant gap remains compared to our method. Crucially, strong performance on held-out evaluators (GPT-5 and Effort, Table 1), which were excluded from Detector-Reward training, validates that RealGen learned generalizable realism rather than merely overfitting to our reward models.

In the arena-style pairwise comparisons, we first analyze the "battle" results against Real images. RealGen demonstrates a significant advantage in this comparison, achieving a win rate approaching 50\%, which suggests its outputs are capable of being confused with reality. In contrast, 8 of the 13 competing models achieved win rates below 30\% against real images, clearly exposing their lack of photorealism. This stark disparity also validates the effectiveness of our Arena-Scoring as a reliable arbiter for realism. Furthermore, Figure~\ref{fig:6} displays the win-rate matrix from the model-vs-model battles. The matrix confirms that RealGen achieves the highest overall win rate, indicating it is consistently selected as the more realistic option when compared directly against its peers.

\begin{table}[t]
    \centering
    \renewcommand{\arraystretch}{1.3}
    \caption{Evaluation on the "Photo" subset of HPDv2 dataset~\cite{wu2023humanpreferencescorev2}. }
    \label{tab:2}
    \vspace{-5pt}
    \resizebox{\linewidth}{!}{
        \begin{tabular}{lcccc}
            \toprule
            \multirow{2}{*}{\textbf{Model}} & \multicolumn{2}{c}{\textbf{Detector-Scoring}} & \multicolumn{2}{c}{\textbf{Aesthetic Scoring}} \\
            \cmidrule(lr){2-3} \cmidrule(lr){4-5}
            & Forensic-Chat & OmniAID & HPSv2.1 & HPSv3 \\
            
            \midrule
            
            \quad FLUX-Pro~\cite{flux2024} & 66.57 & 40.13 & 28.57 & 11.64 \\
            \quad Nano-Banana~\cite{google2025gemini} & 37.44 & 33.09 & 29.37 & 12.23 \\
            \quad GPT-Image-1~\cite{gpt4o} & 63.75 & 27.56 & 29.73 & 12.19 \\
            \midrule
            \quad SDXL~\cite{podell2023sdxl} & 44.53 & 32.22 & 25.92 & 7.27 \\
            \quad FLUX.1 Kontext~\cite{fluxkontext} & 41.25 & 30.66 & 29.06 & 11.43 \\
            \quad Bagel~\cite{bagel} & 37.19& 30.01 & 29.70 & 11.99 \\
            \quad Qwen-Image~\cite{wu2025qwen} & 48.85 & 32.17 & 27.60 & 9.44 \\
            \quad SRPO~\cite{shen2025directly} & 62.65 & 48.23 & 27.88 & 11.11 \\
            \quad FLUX-Krea~\cite{flux1kreadev2025} & 58.00 & 44.96 & 28.76 & 11.39 \\
            
            \rowcolor{myblue}
            \quad \textbf{Ours} & \textbf{71.34} & \textbf{56.93} & \textbf{30.18} & \textbf{13.10} \\
            
            \bottomrule
        \end{tabular}
    }
    \vspace{-5pt}
\end{table}

Fig.~\ref{fig:5} illustrates the qualitative visual comparisons. Base models like FLUX-dev and Bagel tend to produce images with excessive oiliness and unnatural highlights. Qwen-Image often generates overly smooth skin, exhibiting a typical AI artifact. Beyond a distinct "AI plastic" feel, GPT-Image-1 also shows an unnatural warm color cast skewed towards yellow-green. In contrast, the results from RealGen are superior in both texture and detail, appearing visually closer to actual photographs.

To further validate the generalization ability of RealGen, we conducted an additional evaluation on the "photo" subset of HPDv2, with results presented in Table~\ref{tab:2}. The results clearly demonstrate that images synthesized by RealGen not only achieve exceptionally high detector-based realism scores but also rank among the top in human aesthetic preference scores. This provides evidence that our method can significantly enhance photorealism while simultaneously maintaining high aesthetic quality. The supplementary material contains further quantitative and qualitative results, such as detailed artifact analysis from Forensic-Chat and data from human evaluations.

\begin{figure}[t]
    \centering
    \includegraphics[width=1\linewidth]{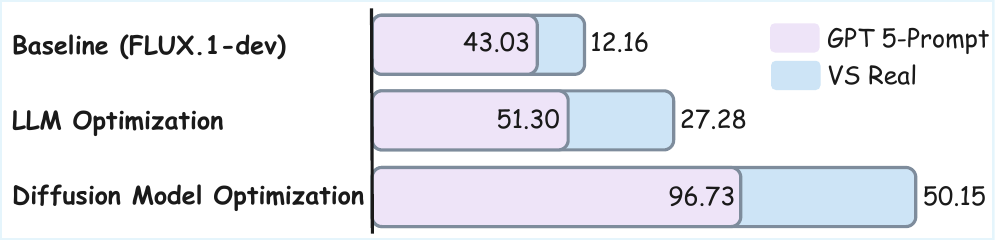}
    \vspace{-15pt}
    \caption{Ablation experiments on different key components.}
    \label{fig:7}
    \vspace{-5pt}
\end{figure}

\begin{figure}[t]
    \centering
    \includegraphics[width=1\linewidth]{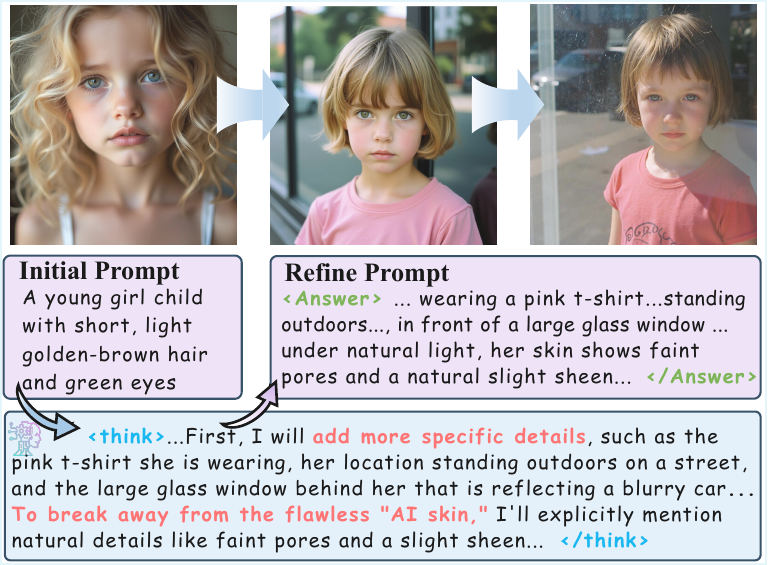}
    \vspace{-10pt}
    \caption{Illustration of the key advantages of synthetic images.}
    \label{fig:8}
    \vspace{-10pt}
\end{figure}

\subsection{Ablation experiments}
\label{subsec:in_context}

As shown in Fig. \ref{fig:7}, we conducted a progressive ablation study to explore the impact of the LLM optimization component and the Diffusion optimization component. For quantitative analysis, we employed two reward-independent metrics: the GPT-prompt Score and vs. Real images Arena-Scoring. The quantitative results indicate that applying only the Phase 1 LLM prompt optimization already leads to more realistic and detailed images by generating richer and more diverse prompts. Building on this, the subsequent inclusion of the Diffusion model optimization further improves image quality. The visualization in Fig. \ref{fig:8} shows the effect of component optimization: the LLM adds realistic descriptions to enrich detail, while optimizing the Diffusion model further enhances the image realism of the person, and glass.

\begin{figure}[t]
    \centering
    \includegraphics[width=1\linewidth]{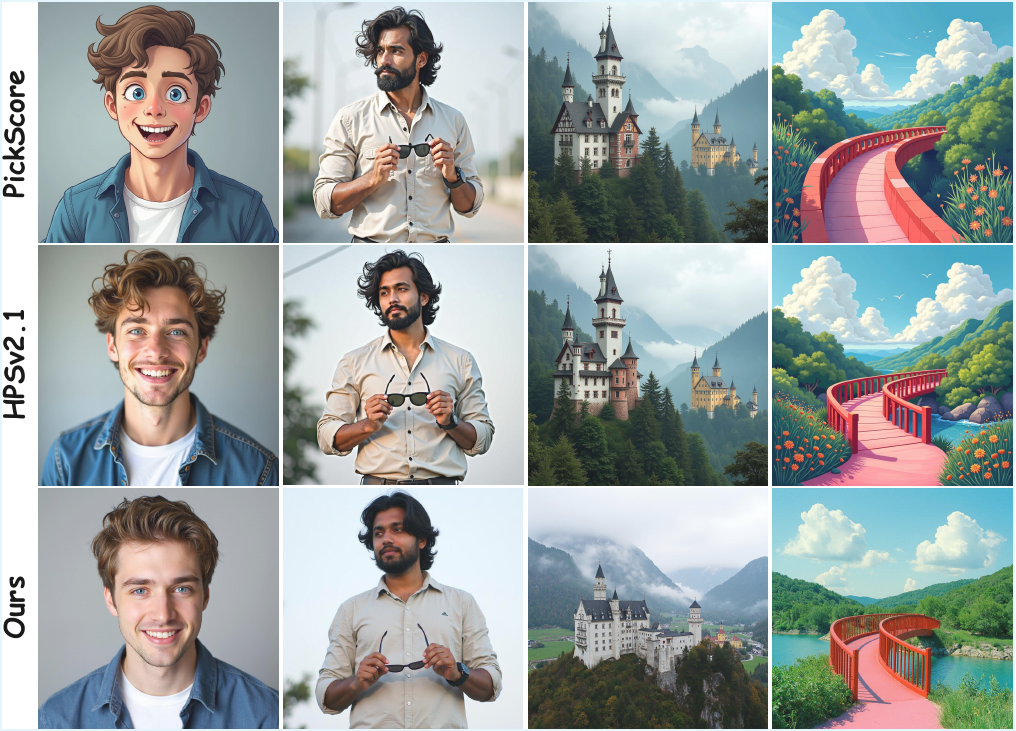}
    \vspace{-15pt}
    \caption{Visualization of the effects of different reward functions.}
    \label{fig:9}
    \vspace{-10pt}
\end{figure}

Furthermore, we explored the distinct impacts of different reward functions on model optimization, comparing human preference rewards against our Detector-score. As shown in Table \ref{tab:3}, when evaluated on multiple metrics held-out from the optimization objective, our Detector-Score demonstrates a clear advantage, proving it guides the model towards superior photorealism. The qualitative results in Figure \ref{fig:9} intuitively explain this gap: reward functions like PickScore and HPSv2.1 tend to bias the model towards cartoonish or artistic styles and still produce "oily" textures on portraits. In contrast, our proposed Detector-Score consistently leads to more realistic and photorealistic results.

\begin{table}[t]
    \centering
    \renewcommand{\arraystretch}{1.3}
    \caption{Ablation experiments on different reward functions.}
    \label{tab:3}
    \vspace{-5pt}
    \resizebox{0.9\linewidth}{!}{
        \begin{tabular}{lcccc}
            \toprule
            \textbf{Method} & Effort & GPT 5 & VS Real & CLIP \\
            
            \midrule
            
            Baseline (FLUX.1-dev) & 14.85 & 43.03 & 12.61 & 86.33 \\
            \quad + PickScore~\cite{kirstain2023pick} & 12.75 & 45.15 & 22.96 & 86.17 \\
            \quad + HPSv2.1~\cite{wu2023humanpreferencescorev2} & 11.46 & 39.00 & 19.12 & 86.24 \\
            \quad + Detector-Score & \textbf{31.71} & \textbf{92.79} & \textbf{43.41} & \textbf{86.80} \\
            \bottomrule
        \end{tabular}
    }
    \vspace{-5pt}
\end{table}

\section{Conclusion}

To address the photorealism gap in current T2I models, we introduced RealGen. Our core contribution is the "Detector Reward" mechanism, inspired by adversarial generation, which utilizes both semantic and feature-level detectors to quantify image realism. By leveraging this reward signal with the GRPO algorithm, we successfully optimized the entire generation pipeline, enhancing image realism and detail; \textbf{this effectively realizes a "Detection for Generation" paradigm in the RL era.} Furthermore, we proposed RealBench, an automated benchmark employing Detector-Scoring and Arena-Scoring, which enables human-free photorealism evaluation. We hope our work will inspire further advancements in photorealistic image synthesis.

{
    \small
    \bibliographystyle{ieeenat_fullname}
    \bibliography{ref}
}

\end{document}